\documentclass{article}
\usepackage{spconf,amsmath,graphicx}
\usepackage{inconsolata}
\usepackage{amsmath,graphicx,amssymb,multirow,setspace}
\usepackage{bm}
\usepackage{booktabs} 
\usepackage{subfigure}
\usepackage[pagebackref,breaklinks,colorlinks]{hyperref}

\title{Tracking Objects and Activities with Attention 
\\for Temporal Sentence Grounding}
\name{Zeyu Xiong$^{1}$\sthanks{Equal contribution. \qquad  $^{\dagger}$ Corresponding author.} \qquad Daizong Liu$^{2*}$ \qquad Pan Zhou$^{1\dagger}$ \qquad Jiahao Zhu$^{1}$}

\address{$^{1}$Hubei Key Laboratory of Distributed System Security, Hubei Engineering \\ Research Center on Big Data Security, School of Cyber Science \\
and Engineering, Huazhong University of Science and Technology \\
$^{2}$Wangxuan Institute of Computer Technology, Peking University \\
 zeyuxiong@hust.edu.cn, dzliu@stu.pku.edu.cn, panzhou@hust.edu.cn, jiahaozhu@hust.edu.cn
}

\begin{document}
%
\maketitle
\begin{abstract}
Temporal sentence grounding (TSG) aims to localize the temporal segment which is semantically aligned with a natural language query in an untrimmed video.
Most existing methods extract frame-grained features or object-grained features by 3D ConvNet or detection network under a conventional TSG framework, failing to capture the subtle differences between frames or to model the spatio-temporal behavior of core persons/objects.
In this paper, we introduce a new perspective to address the TSG task by tracking pivotal objects and activities to learn more fine-grained spatio-temporal behaviors. Specifically, we propose a novel Temporal Sentence Tracking Network (TSTNet), which contains (A) a Cross-modal Targets Generator to generate multi-modal templates and search space, filtering objects and activities, and (B) a Temporal Sentence Tracker to track multi-modal targets for modeling the targets' behavior and to predict query-related segment.
Extensive experiments and comparisons with state-of-the-arts are conducted on challenging benchmarks: Charades-STA and TACoS. And our TSTNet achieves the leading performance with a considerable real-time speed.
\end{abstract}
\begin{figure}[t]
\centering
\includegraphics[width=0.95\columnwidth]{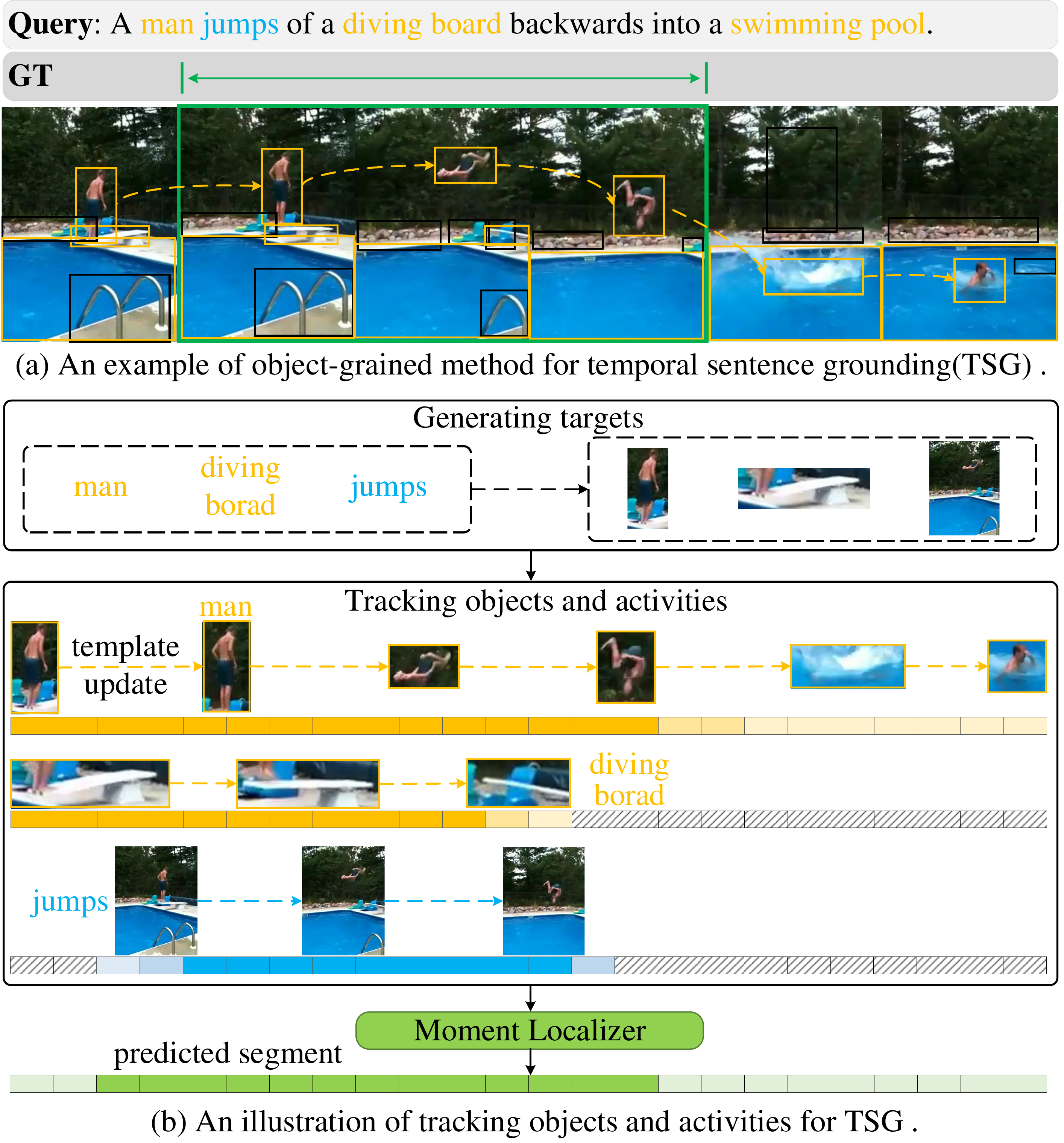}
\vspace{-10pt}
\caption{Illustrations of (a) TSG and (b) tracking-view TSG.}
\label{fig1}
\vspace{-15pt}
\end{figure}

\vspace{-2pt}
\section{Introduction}
\label{sec:intro}
\vspace{-3pt}
Temporal sentence grounding (TSG)~\cite{gao2017tall,anne2017localizing,liu2020jointly} is an important yet challenging task in multi-modal deep learning due to its complexity of multi-modal interactions and complicated context information. As shown in Fig.~\ref{fig1}(a), given an untrimmed video, it aims to determine the segment boundaries including start and end timestamps that contain the interested activity according to a given sentence description.

Most previous works~\cite{gao2017tall,anne2017localizing, liu2021context,liu2022reducing,liu2022skimming,zhu2023rethinking,liu2023hypotheses} first encode and interact the pair of video-query input, and then employ either a proposal-based or a proposal-free grounding head to predict the target segments.
However, these methods extract frame-level video features by a pre-trained 3D ConvNet, which may capture the redundant background appearance in each frame and fails to perceive the subtle differences among video frames with high similarity.
Recently, a few detection-based approaches~\cite{liu2022exploring,xiong2022gaussian,liu2023exploring} have been proposed to capture fine-grained object appearance features inside each frame for focusing more on the foreground contexts. 
Although they achieve promising results, these methods directly correlate all spatial-temporal objects in the entire video through a simple graph- or co-attention mechanism, lacking sufficient reasoning on the most query-specific objects.
\begin{figure*}[t]
\centering
\includegraphics[width=1.0\textwidth]{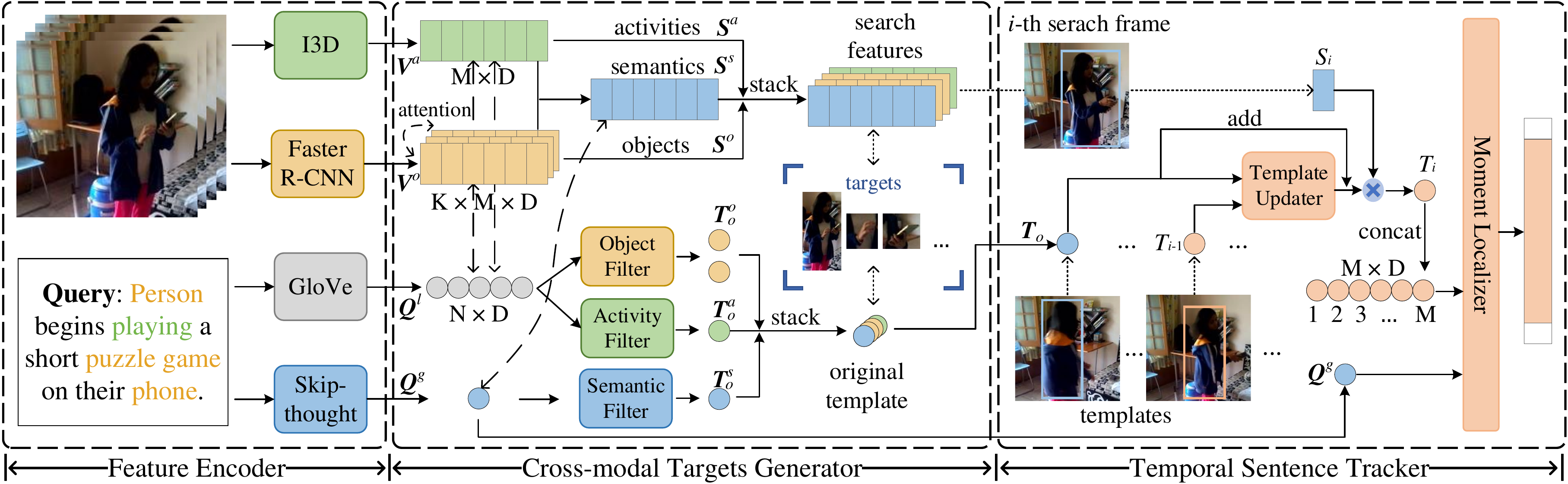}
\vspace{-19pt}
\caption{The overall architecture of our proposed TSTNet. We first encode the video and query to obtain \textbf{V}isual features ${\boldsymbol V^o},\boldsymbol {V}^a$  and \textbf{Q}uery features ${\boldsymbol Q^l,\boldsymbol{Q}^g}$. Then we develop a Cross-modal Targets Generator to generate and filter original \textbf{T}emplates $(\boldsymbol{T}^o_o,\boldsymbol{T}^a_o,\boldsymbol{T}^s_o)$ and \textbf{S}earch space $(\boldsymbol{S}^o,\boldsymbol{S}^a,\boldsymbol{S}^s)$ for latter \textbf{o}bject/\textbf{a}ctivity/\textbf{s}emantics tracking. The Temporal Sentence Tracker is designed for tracking the query-related target corresponding with sentence semantics and predicting the target segment.}

\vspace{-10pt}
\label{fig2}
\end{figure*}

To learn more specific spatial-temporal relations among the extracted objects, as shown in Fig.~\ref{fig1}(b), we adapt the object tracking perspective into the TSG task to correlate the most query-related objects for activity modeling.
Firstly, we generate the multi-modal target templates by selecting core objects and activities (such as "man", "diving broad" and "jumps") and discarding irrelevant ones (black boxes in the video). By aggregating words feature with object-grained visual features, we track the corresponding template in each frame and model the target behavior with continuous target templates. With specific semantics, we deploy a moment localizer to determine the most query-related segment.

However, directly adopting the standard tracking algorithm for TSG will raise two major problems: (1) \textbf{Modal gap}: there is a modal gap between vision and language in TSG while utilizing typical tracking algorithm. (2) \textbf{Ambiguous target}: there is no specific target provided as supervision in TSG, while standard tracking specifies an object to track.
To overcome these two challenges, we develop a novel Temporal Sentence Tracking Network (TSTNet), which contains a Feature Encoder, a Cross-modal Targets Generator and a Temporal Sentence Tracker (Fig.~\ref{fig2}).
Specifically, we first extract the object-grained features and query features by pre-trained detection model, action recognition 3D ConvNet, Glove~\cite{pennington2014glove}, Skip-thought~\cite{kiros2015skip}. Then, we leverage self and co-attention to establish associations among objects, activities and words for bridging the modal gap and generating search space and templates. And we utilize instance filters to further screening the core targets. A dynamic template updater in temporal sentence tracker is to match and dynamically update templates for each frame in search space, modeling the behavior of targets. Finally, we employ a moment localizer to determine the temporal segment and fine-tune the boundaries of it.

Our contributions are summarized as follows: 

(1) We provide a new perspective of tracking objects and activities to address the TSG problem, which can focus more on behavior modeling of core targets;

(2) We propose a novel framework TSTNet, which tackles the differences between TSG and standard tracking;

(3) We demonstrate the predominant effectiveness and efficiency of our TSTNet by evaluating on two benchmarks: Charades-STA and TACoS.



\vspace{-10pt}
\section{Methodology}
\vspace{-3pt}
\label{sec:method}
Given an untrimmed video $\mathcal V$ and a natural language sentence query $\mathcal Q$, the TSG aims at predicting a video segment from time $\tau_s$ to $\tau_e$ corresponding to the same semantic as $\mathcal Q$.

\vspace{-10pt}
\subsection{Feature Encoder}
\vspace{-3pt}
\noindent\textbf{Video Encoder.} In order to model both objects and activities, we extract the appearance-aware and motion-aware features of original videos by the pre-trained Faster R-CNN~\cite{ren2015faster} and C3D/I3D~\cite{carreira2017quo} networks. 

Specifically, for objects, we first uniformly sample fixed $M$ frames from video $\mathcal V$, and then extract $K$ objects from each frame using Faster R-CNN with a ResNet-50 FPN backbone. Therefore, we represent the object features as ${V}^o= \left\{ {o}_{i,j},b_{i,j} \right\} ^{i=M,j=K}_{i=1,j=1}$, where ${o}_{i,j} \in \mathbb{R} ^ {D_o}$ denotes object features with dimension $D_o$, and $b_{i,j} \in \mathbb{R} ^ {4}$ represents the bounding-box coordinate of the $j$-th object in $i$-th frame. 

Since the spatial relationship of instances plays a important role in object behavior modeling, we fuse the spatial information $b_{i,j}$ with object features ${o}_{i,j}$ by \textit{concat} function and Fully Connection (FC), obtaining $\boldsymbol V^o = \{\boldsymbol{v}^{o}_{i,j} \}^{i=M,j=K}_{i=1,j=1}$.

For activity features, we put every 8 frames to a pre-trained 3D ConvNet with stride 4, and sample $M$ output sequence by linear interpolation, which is represented as $\boldsymbol V^a = \left\{\boldsymbol{v}^{a}_{i} \right\}^{i=M}_{i=1}$, where $\boldsymbol{v}^{a}_{i} \in \mathbb{R}^{D}$.

\noindent\textbf{Query Encoder.} For word-level encoding, following previous works~\cite{zhang2019cross,liu2021progressively}, we embed each word in sentence query $\mathcal{Q}$ by Glove~\cite{pennington2014glove}, obtaining the local semantic of every single word: $\boldsymbol{Q}^l = \left\{ \boldsymbol {q}_i \right\} ^{i=N}_{i=0} $, where $\boldsymbol {q}_i \in\mathbb{R} ^ {D}$.
To extract the semantic of the whole sentence, the Skip-thought parser~\cite{kiros2015skip} is employed to capture the global semantic of the whole query, denoted as $\boldsymbol{Q}^g \in\mathbb{R} ^ {D}$.

\vspace{-10pt}
\subsection{Cross-modal Targets Generator}
\vspace{-3pt}
To solve the problem of inconsistency of modality and ambiguous targets to retrieved, we developed a Cross-modal Targets Generator (CTG).

Specifically, we design Search Space Representation (SSR) and Template Generation (TG) across sentences and videos for further cross-modal tracking. As for redundant targets, we develop instance Filters for screening the core semantic-related targets.  

\noindent\textbf{Search Space Representation.}
First, we utilize sentence query guidance to represent video search space for retrieval. Because there exist interactions between objects inside each frame, we first learn the self- and inter-correlation between $K$ objects with attention mechanism~\cite{vaswani2017attention}:
\begin{equation}
\widehat {\boldsymbol{V}}^{o} = \sigma  ({\boldsymbol{V}}^{o}\boldsymbol{W}_1 
({\boldsymbol{V}}^{o}\boldsymbol{W}_2)^\top)
\frac{\boldsymbol{V}^{o}}{{\sqrt D}},
\label{eqself}
\end{equation}
where $\boldsymbol{W}_1,\boldsymbol{W}_2$ are two learnable matrices, $\sigma$ is an activate function. 
Next, we associate words with objects in each frame by leveraging query-guide attention to highlight the word-relevant objects while weakening the irrelevant ones:
\begin{equation}
\begin{aligned}
&\boldsymbol{w}^{q} = \sigma (\frac{1}{{\sqrt D}} (\widehat{\boldsymbol{V}}^{o}\boldsymbol{W}_3) (\boldsymbol{Q}^{l}\boldsymbol{W}_4)^\top),\\
& {\boldsymbol{V}}^{o}_{q} = \boldsymbol{w}^q\boldsymbol{Q}^{l}\boldsymbol{W}_5,
\end{aligned}
\label{eq1}
\end{equation}
where $\boldsymbol{W}_3,\boldsymbol{W}_4,\boldsymbol{W}_5$ are the transform matrices, $\boldsymbol{w}^{q} \in \mathbb{R}^{K \times N}$ represents the correlation between each word-object pair. 
Since too many pre-extracted objects will interfere with tracking and modeling the key targets, we are supposed to filter out redundant backgrounds or instances, and accurately select core objects/activities for tracking.
Therefore, we deploy $k$ adaptive Object Filters to select $k$ core object search space from $K$ object features. In detail, we implement the filter with a linear layer, followed by a Leaky\_ReLU function and a 1d-maxpool layer to activate and filter targets. At last, we obtain $k$ object search space $\boldsymbol {S}^o = \{\boldsymbol _i{S}^o\}^{i=k}_{i=1}$, where $\boldsymbol_
i{S}^o\in \mathbb{R}^{M\times D}$ represents the $i$-th search space for tracking.

For the activity search space representation, similarly, we gain the query-guide activity features $\boldsymbol{V}^{a}_q$ by replacing $\widehat {\boldsymbol{V}}^{o} $ with $\boldsymbol{V}^{a}$ in Eq.~(\ref{eq1}). Then a linear layer followed by a Leaky\_ReLU function is employed to generate the search space $\boldsymbol{S}^{a} \in \mathbb{R}^{M\times D}$.

Considering object or activity alone is not enough to model the semantics and relationships between them. Therefore, we learn the semantic features $\boldsymbol{V}^{s}$  with activity features $\boldsymbol{V}^{a}$ and object features $\widehat {\boldsymbol{V}}^{o} $ by Eq.~(\ref{eq1}), and then calculate element-wise multiplication with $\boldsymbol{Q}^g$ to get semantic search space  $\boldsymbol{S}^{s} \in \mathbb{R}^{M\times D}$


\noindent\textbf{Template Generation.} 
After getting the search space of the video, we need to determine an initial template as the target for matching and tracking. In this case, we consider deeming the sentence query $\boldsymbol{Q}^l,\boldsymbol{Q}^g$ as initial matching template by combining it with instances in videos $\boldsymbol{V}^o,\boldsymbol{V}^a$.

In details, we first calculate the cosine similarity between each word and object at each frame, and obtain the object-aware query feature by equation:
\begin{equation}
\begin{aligned}
&\boldsymbol{w}^s_{i,j,t} = \frac{\boldsymbol{q}_{i,t} (\boldsymbol{v}^{o}_{j,t})^\top}{\Vert \boldsymbol{q}_{i,t}\Vert \Vert \boldsymbol{v}^{o}_{j,t} \Vert},\\
&\widehat{\boldsymbol{q}}_{i,t}^{o} = \sum\limits_{j=1}^{K} \boldsymbol{w}^s_{i,j,t} \boldsymbol{v}^{o}_{j,t},
\widehat{\boldsymbol{Q}}^{o} = \{ \widehat{\boldsymbol{q}}^{o}_{i,t}\} ^{i=N,t=M}_{i=1,t=1},
\end{aligned}
\label{eq2}
\end{equation}
where $\boldsymbol{w}^s_{i,j,t}$ is the similarity between $i$-th word and $j$-th object at frame $t$.

Then, we pick $k$ original object templates $\boldsymbol {T}^o_o=\{_i\boldsymbol {T}^o_o\}^{i=k}_{i=1}\\ \in \mathbb{R}^{k\times D}$ from $\widehat{\boldsymbol{Q}}^{o}$ via $k$ object filters, which corresponds to $k$ search spaces $\boldsymbol {S}^o$.

For activity template generation, similarly, we obtain an activity-aware enhanced query feature  $\widehat{\boldsymbol{Q}}^{a} \in \mathbb{R}^{N \times M \times D}$ by Eq.~(\ref{eq2}) and filter an activity original template  $\boldsymbol {T}^a_o \in \mathbb{R}^{1\times D}$.

The semantic original template $\boldsymbol {T}^s_o \in \mathbb{R}^{1\times D}$ is generated from $ {\boldsymbol{Q}}^{g} $ and ${\boldsymbol{V}}^{s} $ by Eq. (\ref{eq2}) followed by a semantic filter.



\vspace{-10pt}
\subsection{Temporal Sentence Tracker}
\vspace{-5pt}
In order to model the targets' behavior for text-visual alignment, we develop a Dynamic Template Updater (DTU) to track targets in the search space with original templates and then deploy a Moment Localizer to localize the most query-related moment in videos. 

\noindent\textbf{Dynamic Template Updater.} For any search space and template tuple $(\boldsymbol{S},\boldsymbol{T})$, we fuse the original template $ \boldsymbol{T}_o$ and the $(i-1)$-th template  $T_{i-1}$ as a new template for aligning the $i$-th search frame $S_i$, as is shown in Fig.~\ref{fig2} (right). We utilize a template updater $\phi$ to update the templates and then concatenate all templates by sequence, formulated as:
\begin{equation}
\begin{aligned}
&T_i = (\phi(\boldsymbol{T}_o,T_{i-1})+\boldsymbol{T}_o)\cdot S_i,\\
&\boldsymbol{F}_T =[T_1,T_2,...,T_M],
\end{aligned}
\label{eqtrack}
\end{equation}
where $\phi(\cdot)$ is the Feedforward Neural Network (FNN) followed by a GRU~\cite{chung2014empirical} unit. 
$\boldsymbol{F}_T$ contains the behavior information composed of continuous templates.
In practice, we feed different tuples of search space and template into different template updaters, and obtain $\boldsymbol{F}_T^o \in \mathbb{ R}^{k \times M \times D},\boldsymbol{F}_T^a \in \mathbb{ R}^{M \times D},\boldsymbol{F}_T^s \in \mathbb{ R}^{M \times D} $. Then we concatenate them followed by a FC layer and get $\widehat{\boldsymbol{F}}_T$:

Noting that reversed trace of the target also provides rich behavior information, we track the target from the last search frame as Eq. (\ref{eqtrack}) and obtain the reversed features $\widehat{\boldsymbol{F}}_T^{r}$. At last, we concatenate the forward $\widehat{\boldsymbol{F}}_T $ and the reversed $\widehat{\boldsymbol{F}}_T^r $ as $\widetilde{\boldsymbol{F}}_T $. 

\noindent\textbf{Moment Localizer.}
As many temporal localizers are plug-and-play, we follow the previous work~\cite{zhang2019cross} to predict the target moment for fair comparison. 
\vspace{-10pt}
\section{Experiments}
\vspace{-3pt}

\noindent\textbf{Datasets.} \textbf{Charades-STA} dataset was built on Charades by \cite{gao2017tall}, including 9,848 videos of indoor scenarios. By convention, we use 12,408 and 3,720 video-sentence pairs for training and testing. \textbf{TACoS} is collected from the MPII Cooking~\cite{rohrbach2012script}, which contains 127 long videos of cooking scenarios. Following~\cite{gao2017tall}, we obtain 10,146, 4,589 and 4,083 clip-sentence pairs as training ,validation and testing dataset.

\noindent\textbf{Evaluation Metrics.} We adopt "R@$n$, IoU=$\mu$" and "mIoU" metrics for evaluation. The "R@$n$, IoU=$\mu$" denotes the percentage of at least one of top-$n$ predictions having IoU larger than $\mu$. "mIoU" represents the mean average IoU. 

\noindent\textbf{Implementation Details.} For object feature extraction, we utilize Faster R-CNN \cite{ren2015faster} with a ResNet-50 FPN~\cite{lin2017feature} backbone to obtain object features. The number $K$ of extracted objects is set to 15 and the number $k$ of object filter is set to 5. The length of frame sequences $M$ in our model is 64, 200 for Charades-STA and TACoS. For query encoding, we utilize GloVe 840B 300d \cite{pennington2014glove} to embed each word as word features. For model setting, the activate function $\sigma$ is Sigmoid. The hidden dimension $D$ is 512. We sample 800 segment proposals for TACoS and 384 for Charades-STA similar to \cite{zhang2019cross}. We train our model by an Adam optimizer with the learning rate of 0.0008 for 60 epoches. Batch size is 64.

\begin{table}[t]
    \small
    \centering
	\setlength{\tabcolsep}{7.0 pt}
	\begin{tabular}{l |c| c c c | c}
		\toprule
		\multicolumn{6}{c}{Charades-STA} \\
		\hline
		\multirow{2}{*}{Methods} & \multirow{2}{*}{Feature}&\multicolumn{3}{c |}{$\text{R@}1, \text{IoU}=$} & \multirow{2}{*}{mIoU} \\
          && $0.3$ & $0.5$ & $0.7$ & \\
        \hline

        CBP     &C3D& -     & 36.80 & 18.87 & 35.74  \\
        2DTAN  &VGG& -     & 39.81 & 23.31 & -     \\
        VSLNet  &I3D& 70.46 & 54.19 & 35.22 & {50.02} \\
        LGI     &I3D& {72.96} & {59.46} & {35.48} & {51.38}     \\ 
        CPN     &I3D& 75.53 & 59.77 & 36.67 & 53.14\\
        IA-Net  &I3D& -     & 61.29 & 37.91 & -    \\
        DRFT    &I3D+F+D& \textit{76.68} & {63.03} & {40.15} & \textit{54.89} \\
        MARN   &I3D+Obj & -  &\textit{66.43} & \textit{44.80}   & -  \\
        \hline
    	\multirow{2}{*}{\textbf{TSTNet}} &C3D+Obj& {76.26} & {65.34} & {43.61} & {56.76} \\
          &I3D+Obj& \textbf{77.62} & \textbf{67.49} & \textbf{45.21} & \textbf{57.82} \\
        \toprule
	\end{tabular}
	\vspace{-5pt}
	\caption{Comparison with SOTAs on Charades-STA.}
	\label{tab:sota_charades}
\end{table}


\begin{table}[t]
    \small
	\centering
	\setlength{\tabcolsep}{7.0 pt}
	\begin{tabular}{l|c | c c c | c}
		\toprule
		\multicolumn{6}{c}{TACoS}\\
        \hline
    
		\multirow{2}{*}{Methods} & \multirow{2}{*}{Feature}&\multicolumn{3}{c |}{$\text{R@}1, \text{IoU}=$} & \multirow{2}{*}{mIoU} \\
          && $0.3$ & $0.5$ & $0.7$ & \\
        \hline
        CMIN   &C3D&24.64 & 18.05 & -     & - \\

        CBP    &C3D &27.31 & {24.79} & 19.10 & 21.59 \\
        2DTAN & C3D&37.29 & 25.32 & -     & -     \\
        VSLNet &I3D&{29.61} & 24.27 & {20.03} & {24.11} \\

        IA-Net &I3D &37.91 & 26.27 & -     & -     \\
        CPN    &I3D &48.29 & 36.58 & \textit{21.25} & \textit{34.63} \\
        MARN   &I3D+Obj &\textit{48.47} & \textit{37.25}& -     & -  \\
        \hline
        \multirow{2}{*}{\textbf{TSTNet}} &C3D+Obj & 50.21 & 38.47 & 23.12 & 35.26 \\
         &I3D+Obj & \textbf{53.39} & \textbf{41.23} & \textbf{26.62} & \textbf{37.83} \\
        \toprule
	\end{tabular}
	\vspace{-5pt}
	\caption{ Comparison with SOTAs on TACoS.}
    \vspace{-10pt}
	\label{tab:sota_tacos}
\end{table}
\begin{table}[t]
    \small
	\centering
	\setlength{\tabcolsep}{5pt}
	\begin{tabular}{c| c c c  c}
	    \toprule
         Methods & TGN & 2DTAN & CMIN  & TSTNet\\
        \hline
        V-QPS & 2.23 & 3.89 & 86.29 & \textbf{103.27}  \\
        Parameters & 166  & 363 & 78 &  \textbf{67}\\ 
        Accuracy & 18.89 & 25.32 & 18.05  & \textbf{41.23} \\
        \toprule
	\end{tabular}
 	\vspace{-5pt}
	\caption{Effyciency comparision in terms of video-query pairs per second (V-QPS), Parameters (Mb) and Accuracy (R@1, IoU=0.5 metric) on TACoS dataset.}
	
	\label{tab:efficiency}
\end{table}

\vspace{-10pt}
\subsection{Experimental Results and Analysis}
\vspace{-3pt}
We compare the proposed TSTNet with the following state-of-the-arts: (1) \textit{Proposal-based} methods:  CBP~\cite{Wang2020TemporallyGL}, 2DTAN \cite{zhang2019learning}, CMIN~\cite{zhang2019cross},  IA-Net~\cite{liu2021progressively}; (2) \textit{Proposal-free} methods:  VSLNet~\cite{zhang2020span}, LGI~\cite{mun2020local} , CPN~\cite{zhao2021cascaded}; (3) \textit{Multi-stream} methods: DRFT~\cite{chen2021end}, MARN~\cite{liu2022exploring}. The best results are in \textbf{bold} and the second bests are in \textit{italic}. 

\noindent\textbf{Quantitative Comparion.}
As summarized in Table~\ref{tab:sota_charades} and \ref{tab:sota_tacos}, our proposed TSTNet surpasses all existing methods on two datasets. Observe that the performance improvements of TSTNet are more significant under more strict metrics (R@1, IoU=0.7), indicating that TSTNet can predict more precise moment boundaries of untrimmed videos. 
As multi-stream methods, DRFT integrates the three modalities of visual information RGB~(I3D), optical flow~(F) and depth maps~(D), and MARN fuses activity (I3D) with object (Obj). They perform well and imply that combining multi-dimension sources helps the model learn more accurate semantics. Differing from DRFT and MARN, our TSTNet traces objects from a finer granularity and mines more explicit behavior of core targets, thus outperforming better in results.

\noindent\textbf{Efficiency Comparison.} We compare the inference speed and effectiveness of our TSTNet with previous methods on a single Nvidia Quadro RTX5000 GPU on TACoS dataset. Table \ref{tab:efficiency} shows that TSTNet achieves a significantly faster inference speed and a lightweight model size. 

\begin{table}[t]
    \small
	\centering
	\setlength{\tabcolsep}{7.0 pt}
	\begin{tabular}{c|c| c c c}
		\toprule
		\multirow{2}{*}{Components} & \multirow{2}{*}{Changes}&\multicolumn{3}{c }{$\text{R@}1, \text{IoU}=$} \\
         & & $0.3$ & $0.5$ & $0.7$ \\
        \hline
       Cross-modal  &w/o SSR  & 72.59 & 62.21 & 39.79 \\
        Target&w/o TG   & 74.25 & 63.82 & 41.59 \\
        Generator&w/o Filter& 73.12 & 62.74 & 40.27  \\  
        \hline
        
        \multirow{3}{0.1\textwidth}{\centering Temporal Sentence Tracker} &w/o  DTU  & 70.12 & 60.49 & 38.84 \\
         &w/o  GRU  & 72.31 & 61.92 & 39.65 \\
        &w/o  $\widehat{\boldsymbol{F}}_T^{r}$  & 73.61 & 62.64 & 40.35 \\
        \hline
        \multicolumn{2}{c|}{\textbf{Full}}   & \textbf{77.62} & \textbf{67.49} & \textbf{45.21} \\
        \toprule
	\end{tabular}
 	\vspace{-3pt}
	\caption{Ablation study on Charades-STA dataset.}
	\vspace{-10pt}
	\label{tab:main_ablation}
\end{table}

\noindent\textbf{Ablation Study.} As shown in Table \ref{tab:main_ablation}, we verify the contribution of several modules in our TSTNet: we remove SSR, TG, Filter in Cross-modal Target Generator, and DTU, GRU, reversed feature $\widehat{\boldsymbol{F}}_T^{r}$ in Temporal Sentence Tracker (mentioned in Sec.~\ref{sec:method}). The result manifests each above component provides a positive contribution.
\vspace{-10pt}
\section{Conclusion}
\vspace{-3pt}
In this work, we solve the TSG task with a multi-modal instance tracking framework and propose the TSTNet. With this effective and efficient framework, TSTNet outperforms state-of-the-arts on two challenging benchmarks.

\noindent \textbf{Acknowledgments.} This work was supported by National Natural Science Foundation of China under No. 61972448.

\vfill\pagebreak

\bibliographystyle{IEEEbib}
\bibliography{custom}

\end{document}